%% file: paper.tex
\documentclass[submission,copyright,creativecommons,noderivs,noncommercial]{eptcs}
\providecommand{\event}{7th International Workshop on\\ Symbolic-Numeric Methods for\\ Reasoning about CPS and IoT} % Name of the event you are submitting to
\usepackage{url}
\usepackage{microtype}
\usepackage{graphicx}
\usepackage{tikz,pgfplots}
\usepackage{xcolor}
\usepackage{wrapfig}

\usepackage{float}
\usepackage{subcaption}
\usepackage{placeins}
\usepackage{nicefrac}

\usepackage{tabularx}
\usepackage{booktabs}
\usepackage{array}
\newcolumntype{R}{>{\raggedleft\arraybackslash}X}
\newcolumntype{C}{>{\centering\arraybackslash}X}

\usepackage{amsmath}
\usepackage{amssymb}

\usepackage{siunitx}

\usepackage[ruled,vlined]{algorithm2e}

\usepackage{csquotes}

\usepackage{paralist}



\newcommand{\Autoref}[1]{%
	\begingroup%
	\renewcommand\equationautorefname{Equation}%
	\renewcommand\figureautorefname{Figure}%
	\renewcommand\subsectionautorefname{Section}%
	\renewcommand\subsubsectionautorefname{Section}%
	\renewcommand\sectionautorefname{Section}%
	\renewcommand\tableautorefname{Table}%
	\autoref{#1}%
	\endgroup%
}

\def\subsubsectionautorefname{Sec.}
\def\subsectionautorefname{Sec.}
\def\sectionautorefname{Sec.}
\def\tableautorefname{Tab.}
\def\figureautorefname{Fig.}
\def\equationautorefname{Eq.}

\usepackage{todonotes}

\usepackage[usestackEOL]{stackengine}
\usepackage[inline]{enumitem}

\DeclareMathOperator{\sig}{sig}

\newcommand*{\inp}{\ensuremath{\mathrm{in}}}
\newcommand*{\outp}{\ensuremath{\mathrm{out}}}

\title{Verification of Sigmoidal Artificial Neural Networks using iSAT}

\author{
	Dominik Grundt, Sorin Liviu Jurj,\\ Willem Hagemann
	\institute{German Aerospace Center e.V (DLR)\\ Institute of Systems Engineering for Future Mobility\\ Oldenburg, Germany}
	\email{\Longunderstack{dominik.grundt@dlr.de, sorin.jurj@dlr.de,\\willem.hagemann@dlr.de}}
	\and
	\Longunderstack{Paul Kr\"oger, Martin Fr\"anzle}\smallskip
	\institute{Carl von Ossietzky University Oldenburg\\ Oldenburg, Germany}
	\email{\Longunderstack{paul.kroeger@uol.de,\\ martin.fraenzle@uol.de}}
}

\def\titlerunning{Verification of Sigmoidal Artificial Neural Networks using iSAT}
\def\authorrunning{D. Grundt, S.L. Jurj, W. Hagemann, P. Kr\"oger, M. Fr\"anzle}

\pgfplotsset{compat=1.16}
\begin{document}

\maketitle

\input{further_input/abstract}

\thanks{\small{\noindent%
		This work has received funding from the German
		Federal Ministry of Economic Affairs and Climate Action (BMWK) through the KI-Wissen
		project under grant agreement No. 19A20020M, and from the State of Lower Saxony
		within the framework ``Zukunftslabor Mobilit{\"a}t Niedersachsen''
		(\url{https://www.zdin.de/zukunftslabore/mobilitaet}).}}

\input{further_input/introduction}

\input{further_input/related_work}
\input{further_input/approach}

\input{further_input/experiment_setup/setup}

\input{further_input/feasibility}
%\section{Preprocessing Patch}
\input{further_input/prep_patch.tex}
\input{further_input/new_benchm.tex}
\input{further_input/discussion.tex}

\bibliographystyle{eptcs}
\bibliography{lib}
\end{document}

%% file: further_input/abstract.tex
\begin{abstract}
    This paper presents an approach for verifying the behaviour of nonlinear
    Artificial Neural Networks (ANNs) found in cyber-physical safety-critical
    systems. We implement a dedicated interval constraint propagator for the
    sigmoid function into the SMT solver iSAT and compare this approach with a
    compositional approach encoding the sigmoid function by basic arithmetic
    features available in iSAT and an approximating approach. Our experimental
    results show that the dedicated and the compositional approach clearly
    outperform the approximating approach.  Throughout all our benchmarks, the
    dedicated approach showed an equal or better performance compared to the
    compositional approach.
\end{abstract}

%% file: further_input/introduction.tex
\section{Introduction}

In the age of highly automated systems and the development of autonomous
systems, a possible application scenario for ANNs is to use them as controllers
for safety-critical cyber-physical systems (CPSes) \cite{verisig}. Such CPSes
capture the often complex environment, analyse the data and make control
decisions about the future system behaviour. Guarantees on compliance with safety
requirements, e.g., that human lives are not endangered, are of utmost
importance. Whenever such guarantees are obtained via formal verification of
the system behaviour, an ANN being a component of the system under analysis has
also to be subject to verification~\cite{verifisurv}.

The underlying weighted summation of the input neurons before application of the
activation function can be represented by simple linear combinations and is
therefore very appealing for classical verification methods dealing with linear
arithmetic.  However, this observation is deceptive when nonlinear activation
functions are part of the ANN as such nonlinear functions are often hard to
analyse in themselves~\cite{verifisurv}. Apart from restricted decidability
results for reachability problems as in~\cite{verisig}, the runtime of
algorithms for automatic verification suffers from the multiple occurrence of
nonlinear activation functions in complex ANNs such that only relatively small
networks could be tackled. Depending on the class of activation functions and
the underlying possibly necessary abstractions thereof, recent approaches were
able to deal with ANNs comprising 20 to 300
nodes~\cite{katz2017reluplex,hysatforabstr,pulina:smt-nn:2012}.

So-called satisfiability modulo theory (SMT) solvers implement algorithms that
search for solutions to Boolean combinations of arithmetic constraints or prove
the absence thereof. Such SMT solvers are often used for automatic verification
of safety-critical properties in CPSes~\cite{Scheibler2015,katz2017reluplex}.
Here, a symbolic description of the possible system behaviour in terms of a
constraint system is analysed for the existence of a system run reaching a state
violating a safety requirement.
In order to apply automatic verification to ANNs with nonlinear activation
functions, which is necessary for automatic verification of CPSes comprising
such ANNs, the SMT solver used must support the corresponding class of
arithmetic.

The SMT solver iSAT~\cite{isat,herde} solver reasons over the\,---\,in general
undecidable\,---\,theory of Boolean combinations of nonlinear arithmetic
constraints. It tightly couples the well-known Boolean decision procedure
DPLL~\cite{davis} with interval constraint propagation (ICP)
\cite{rossi2006handbook} for real arithmetic. In addition to basic arithmetic
operations, iSAT also supports nonlinear and transcendental operators. Given an
input formula and interval bounds of the real-valued variables, iSAT searches
for a satisfiable solution within the interval bounds.
The iSAT algorithm incorporates an alternation of deduction and decision steps.
A decision step selects a variable, splits its current interval and decides for
one of the resulting intervals to be the search space of the variable in the
further course of the search. A deduction step then applies forward and backward
interval constraint propagation until either a fixed point is reached, or an
empty interval is derived which means that the formula is unsatisfiable under
the current assumptions (decisions). The latter result causes a revoke of
deductions of interval bounds and a reversal of decisions which yields, if
possible, a decision for a yet unexplored part of the search space. If no
decision for an unexplored part of the search space is possible, the formula is
proven to be unsatisfiable under the initial assumption of the bounds on the
search space.

Due to the inherent availability of nonlinear operators, iSAT seems to be a
promising tool for verification of ANNs employing nonlinear activation
functions. ANNs such as Deep Learning (DL) architectures make use of the
nonlinear and transcendental sigmoid function (and their evolved variations) in
hidden and output layers \cite{actfunctions}.  Although the solver iSAT can
already represent the sigmoid function by a composition of existing propagators,
this paper pursues the hypothesis that for nonlinear functions, a propagator
which can exactly propagate the function in once (in following called dedicated
propagator) is preferable. For the investigation of this hypothesis, the sigmoid
function is used in this paper.  We therefore present an approach for verifying
sigmoidal ANNs based on a dedicated interval constraint propagator for the
sigmoid function integrated into iSAT. In our experiments, we demonstrate based
on runtime comparisons that our approach points into a promising direction for
future research on the verification of nonlinear ANNs.

The paper is organised as follows. In \autoref{sec:relatedwork}, we present the
related work regarding different approaches for verifying ANNs.
\Autoref{sec:proposedapproach} details the proposed approach for verifying ANNs
using the iSAT solver and is divided into several parts.  We first discuss and
compare different encoding schemes for the sigmoid activation function
(\autoref{sec:translation}), present our experimental setup
(\autoref{sec:experimentalsetup}), and present some preliminary results
regarding the encoding schemes (\autoref{sec:firstexperiments}). Afterwards, we
report on improvements in the preprocessing step of iSAT
(\autoref{sec:preprocessing}) that proved helpful for the following verification
of severe safety properties that is discussed in
\autoref{sec:verificationexperiments}.  Finally, in \autoref{sec:conclusions} we
present the conclusions and future work of this paper.

%% file: further_input/related_work.tex
\section{Related Work}
\label{sec:relatedwork}

In the literature, there is a high research interest regarding the verification
of ANNs, especially regarding cyber-physical safety-critical systems.

The authors in \cite{hysatforabstr} present an approach that verifies the safety
of a fully connected feedforward network and applies automatic corrections to
the neurons. For this purpose, an industrial manipulator with its kinematic
properties was used as a physical system. More exactly, an ANN with the sigmoid
function being used as an activation function was used to predict the final
position depending on the given joint angles of the industrial manipulator. The
presented approach deals with an abstraction of the concrete ANN, were the
the abstraction domain was chosen to be the set of (closed) intervals of
real numbers. The corresponding abstraction of the sigmoid
activation function is hence given by piecewise interval boxes
of fixed width that encapsulate the concrete function.
The chosen abstraction is consistent, i.e,
if the abstraction of a safety property is valid in the abstract domain,
then it must hold also for the concrete domain. The verification task
could be handled by the SMT solver HySAT\footnote{HySAT is a predecessor version of iSAT.}. When HySAT produces an abstract counterexample, it has to
be analysed whether it has a concretisation or is spurious.
In the later case, a refinement of the interval boxes is triggered.
In addition, the authors propose to use spurious counterexamples for a
'repair' of the ANN, i.e., they propose to add a correct labelled concretisation
to the training set for later trainings. The corresponding tool is called NeVer.

In 2017, the authors in \cite{katz2017reluplex} from Stanford University
published the algorithm Reluplex.  Reluplex is based on the well-known Simplex
algorithm for linear programming. The idea of Reluplex is to extend
the simplex algorithm with new rules for the piecewise linear ReLU-activation function (Rectified Linear Unit).
The task was to verify a prototypical DNN implementation of the new collision
avoidance system ACAS Xu \cite{acas} for unmanned aerial vehicles (UAVs).
The deep neural networks (fully connected and
feed-forward) were trained on a look-up table that holds information about which
manoeuvre advisories are to be executed next depending on seven input parameters.
The evaluation was performed on 45 networks with 300 hidden neurons each, where
the Reluplex algorithm was able to prove (and disprove)
that certain safety-critical properties hold for the given DNN. 
Furthermore, the robustness of the deep neural network system was
tested. It was possible to determine individual input neurons and their
sensitivity to perturbations by adding a perturbation factor to the input and
checking the output \cite{katz2017reluplex}.

The Reluplex approach has been included into the Marabou framework \cite{marabou} in 2019. Marabou works with ANNs using arbitrary piecewise-linear activation
functions and follows the spirit of Reluplex as it uses a lazy SMT-technique
that postpones a case-split of nonlinear constraints as far as possible,
in the hope that many of them prove irrelevant for the property
under investigation.

Also in 2017, another approach to verification of deep neural networks
was published in \cite{oxford}.  Here, the safety analysis
of classification decisions is driven by so-called
adversarials. For a given input vector $x$, the invariance of
the classification decision of the deep neural network is to be
proven or disproven layer by layer by considering disturbances or manipulations
within a region around the layer's activation. Hence, a deep neural network is
safe if, for a given input vector and region, the application of specific
manipulations does not lead to a change in the classification decision.
Such a region can be spanned manually or automatically, by taking a
hyperrectangle around the activation. The hyperrectangle has a nontrivial
extent in a given number of dimensions, these being the dimensions where the
activation value is furthest away from the average activation value of the layer.
Manipulations for the selected region can also be determined manually or by adding or subtracting a small numerical value (so-called small span) depending on the positive or negative deviation of the average activation. After the selection of the region and the manipulations, the SMT Solver Z3 is used to check whether a misclassification is already possible in the considered layer due to the applied manipulations compared to the input. If there is no misclassification, an automatic determination of the region and manipulations in the next layer depends on the region and manipulations of the previously considered layer. In order to automatically find valid manipulations, the solver Z3 is also used. Thus, a safety statement can be made whether the classification network is robust on the specific input against perturbations within the initially specified region or not.
Focusing on classification problems, evaluations on image classification deep
neural networks were performed in this
publication. These include a neural network for handwritten digit recognition
(greyscale) based on the MNIST data-set \cite{mnist_dataset}, CIFAR-10 for object
recognition \cite{cifar10}, and ImageNet \cite{imagenet} for 3-channel colour
images \cite{oxford}. The evaluation showed that the approach scales well in terms of neural network size (state-of-the-art 16-layers image classification networks) and the amount of total parameters (up to 138,357,544 in experiments) since each layer is checked individually for misclassifications. Unfortunately, no temporal information was provided for the performed experiments. A statement about the safety of a neural network is made by falsification. If no misclassification was found, the network is safe considering the specific vector, the selected region, and the set of
manipulations. The evaluation was done with the solver Z3 \cite{demouraz3,
oxford}.

In 2018, the authors of \cite{ai2} presented the
verification of safety and robustness properties of feed-forward neural networks
with max-pooling layers as well as convolutional neural networks.  The presented
approach uses the classical framework of abstract interpretation to
overapproximate the behaviour of the original neural network. For the
approximation the authors investigated different domains of polyhedral shapes,
e.g., boxes, zonotopes (restricted class of point-symmetrical polyhedra) and
general convex polyhedra. The approach was evaluated on fully-connected
feedforward neural networks with ReLU-function as activation function trained
with the MNIST \cite{mnist_dataset} and CIFAR \cite{cifar10} data-set. The
results have shown better scalability and precision on state-of-the-art neural
networks compared to Reluplex \cite{katz2017reluplex}. The state-of-the-art
neural network analyser ERAN \cite{eran} subsumes the presented approach.   

In this work, we present the approach
of verifying fully connected feedforward neural networks containing sigmoid
activation functions using the SMT solver iSAT \cite{isat} that builds on 
interval constraint propagation and supports nonlinear and transcendental
arithmetic.

%% file: further_input/approach.tex
\section{Verifying Sigmoidal ANNs with iSAT}
\label{sec:proposedapproach}

In this section, we present our approach regarding verification of ANNs using
the SMT solver iSAT. In \autoref{sec:translation} we specify how the ANNs and
their desired safety properties are encoded into iSAT. We describe three
different approaches to encode the sigmoid function into iSAT.  In
\autoref{sec:experimentalsetup} we introduce our benchmark set including the
used verification targets. A mild subset of the verification targets was used
for a first comparison of the encoding approaches as presented in
\autoref{sec:firstexperiments}. As in this early phase of our investigation it
turned out already, that iSAT has some runtime issues with regard to the
preprocessing of the input formula, we report on our countermeasures to
improvements the preprocessing in \autoref{sec:preprocessing}. Finally, in
\autoref{sec:verificationexperiments} we report on our experience on safety
verification using a subset of severe verification targets.

\subsection{Translating ANN into iSAT}\label{sec:translation}

\paragraph{Safety Properties}

An ANN encodes a functional relation $o=F(i)$ between the network input $i$ and
the network output $o$.  Since all edges between neurons, weights and transfer
functions are known for a trained network, this information can be extracted,
and $F$ can be written as a comprehensive arithmetic expression, and hence fed
into the SMT solver iSAT.  A \emph{safety property} for the ANN relates input
constrains $\Phi(i)$ with output constraints $\Psi(o)$\,---\,both given again as
arithmetic expressions\,---\,, and we say that the ANN satisfies the safety
property if and only if the implication $(o=F(i) \land \Phi(i)) \to \Psi(o)$
holds for all inputs $i$ and outputs $o$.  That is, whenever iSAT certifies that
the negation of the safety property, $o=F(i) \land \Phi(i) \land \lnot\Psi(o)$,
is unsatisfiable, we have proven the safety property for the ANN. On the other
hand, if iSAT provides a solution to the given problem it found a counterexample
for the desired property. Often, iSAT provides a so-called candidate solution
only. A candidate solution specifies for each solver variable an interval in
which a possible solution could be found. Such a candidate solution needs to be
analysed in subsequent steps.

\paragraph{Sigmoid as Activation Function}
As already said in the introduction we aim at the verification of ANNs using
the sigmoid function 
\begin{align}
    sig(x) : \mathbb{R} \to (0, 1) \;\text{, }\;\;
    x \mapsto \frac{1}{1 + \mathrm e^{-x}} \;\text{, }\;\; \label{eq:sig}
\end{align}
as activation function, and compare different encoding approaches of the
sigmoid function into iSAT.

\begin{figure}[hb!]
  \begin{minipage}[t]{0.31\textwidth}
    \centering
    \resizebox{\textwidth}!{
      \pgfplotsset{compat=1.16}
      \begin{tikzpicture}
        \begin{axis}[
            axis x line=middle, %position der x-achse bottom, middle, top, none
            axis y line=middle, %position der y-achse right, middle, left, none
            xlabel={$x$}, %x-achsenbeschriftung
            ylabel={$y$}, %y-achsenbeschriftung
            xtick={-6,-5,...,5,6}, % x-achsenskalierung
            ytick={0,0.1,0.2,...,0.8,0.9,1.0}, % y-achsenskalierung
            xmin=-7,  % definitionsbereich minimumwert
            xmax=7,   % definitionsbereich maximumwert
            ymin=-0.1,  % wertebereich minimumwert
            ymax=1.1,   % wertebereich maximumwert
            label style={font=\tiny},
            tick label style={font=\tiny,color=gray}  
          ]
          \addplot[domain=-6.2:6.2, smooth, variable=\x, blue]
          plot ({\x}, {1/(1+e^(-\x)})
          node[anchor=north east, inner ysep=2ex, black]{$y=\sig(x)$};
        \end{axis}
      \end{tikzpicture}
    }
    \caption{The graph of the sigmoid function $y=\sig(x)$.}\label{fig:sig_pure}
  \end{minipage}
  \hfill
  \begin{minipage}[t]{0.31\textwidth}
    \centering
    \resizebox{\textwidth}!{
      \pgfplotsset{compat=1.16}
      \begin{tikzpicture}
        \begin{axis}[
            axis x line=middle, %position der x-achse bottom, middle, top, none
            axis y line=middle, %position der y-achse right, middle, left, none
            xlabel={$x$}, %x-achsenbeschriftung
            ylabel={$y$}, %y-achsenbeschriftung
            xtick={-6,-5,...,5,6}, % x-achsenskalierung
            ytick={0,0.1,0.2,...,0.8,0.9,1.0}, % y-achsenskalierung
            xmin=-7,  % definitionsbereich minimumwert
            xmax=7,   % definitionsbereich maximumwert
            ymin=-0.1,  % wertebereich minimumwert
            ymax=1.1,   % wertebereich maximumwert
            label style={font=\tiny},
            tick label style={font=\tiny,color=gray}  
          ]
          \addplot[domain=-6.2:6.2, smooth, variable=\x, blue]
          plot ({\x}, {1/(1+e^(-\x)})
          node[anchor=north east, inner ysep=2ex, black]{$y=\sig(x)$};
          \foreach \x in {-6,-5.5,-5,-4.5,-4,...,5.5,6} {
            \edef\temp{\noexpand\draw[fill=green, fill opacity=0.2] (\x,{1/(1+e^-\x}) rectangle (\x+0.5,{1/(1+e^-(\x+0.5)});}
            \temp
          }
          \draw[fill=green, fill opacity=0.2] (-6.9,0) rectangle (-6,{1/(1+e^-(-6))});
          \draw[fill=green, fill opacity=0.2] (6.9,1) rectangle (6,{1/(1+e^-6)});
        \end{axis}
    \end{tikzpicture}}
    \caption{Approximation of $y=\sig(x)$ by encapsulating interval boxes.}
    \label{fig:sig_approx}
  \end{minipage}
  \hfill
  \begin{minipage}[t]{0.31\textwidth}
    \centering
    \resizebox{\textwidth}!{
      \pgfplotsset{compat=1.16}
      \begin{tikzpicture}
        \begin{axis}[
            axis x line=middle, %position der x-achse bottom, middle, top, none
            axis y line=middle, %position der y-achse right, middle, left, none
            xlabel={$x$}, %x-achsenbeschriftung
            ylabel={$y$}, %y-achsenbeschriftung
            xtick={-6,-5,...,5,6}, % x-achsenskalierung
            ytick={0,0.1,0.2,...,0.8,0.9,1.0}, % y-achsenskalierung
            xmin=-7,  % definitionsbereich minimumwert
            xmax=7,   % definitionsbereich maximumwert
            ymin=-0.1,  % wertebereich minimumwert
            ymax=1.1,   % wertebereich maximumwert
            label style={font=\tiny},
            tick label style={font=\tiny,color=gray}  
          ]
          \addplot[domain=-6.2:6.2, smooth, variable=\x, blue]
          plot ({\x}, {1/(1+e^(-\x)})
          node[anchor=north east, inner ysep=2ex, black]{$y=\sig(x)$};
          \draw[dashed] (1.3,0.08) rectangle (-5.8,0.97)
          node[anchor=north west] {$X\times Y$};
          \draw[dotted] (1.3,0.08) rectangle ({ln(0.08/(1-0.08))},{1/(1+e^-1.3)})
          node[anchor=north east] {$X'\times Y'$};
        \end{axis}
      \end{tikzpicture}
    }
    \caption{\hfill%
      Interval constraint propagation for $\sig(x)$. $X'$ and $Y'$ are the smallest intervals that still contain all solutions of $y=\sig(x)$ that were previously in $X$ and $Y$.}
    \label{fig:sig}
  \end{minipage}
\end{figure}

\paragraph{iSAT Encoding Approaches}
In a first step, we were mainly interested in a comparison of 
different approaches for the arithmetic encoding of the
functional relation $o=F(i)$ with regard to the sigmoid activation function.
\begin{enumerate}[label={\alph*})]
\item As iSAT provides the necessary arithmetic means to express 
  $\sig(x)$ as a composition of basic arithmetic operations
  and the exponential function as indicated on the right hand side of
  \autoref{eq:sig}\footnote{Since iSAT does not support division directly,
  the expression $z=\frac{1}{1 + \mathrm e^{-x}}$ is encoded
  as $1 = z \cdot (1 + \exp(-x))$.}, this yields the
  canonical \emph{compositional approach}.
\item Since we assumed that a specialised propagator for sigmoid would behave better
  for the verification task, we built such an specialised propagator into iSAT, which led to the \emph{dedicated approach}.
\item Finally, we were also interested in an
  \emph{approximating approach} that uses\,---\,similar to the sigmoid
  abstraction in \cite{hysatforabstr}\,---\,a piecewise interval encapsulation
  of the sigmoid function, cf. \autoref{fig:sig_approx}.
  Within the domain interval $[-8,8)$ we encapsulated the sigmoid function
  in interval boxes of a fixed width $p=0.5$ and for the remaining $x$-values
  below $-8$ and above $8$ we used unbounded interval boxes. Hence, we replaced
  any occurrence of the relation $y=\sig(x)$ by the formula
  \begin{gather*}
    \Big( x \in \big(-\infty, 8\big) \,\to\, y \in \big[ 0, \sig(-8) \big)  \Big) \land
    \Big( x \in \big[ -8, -7.5 \big) \,\to\, y \in \big[ \sig(-8), \sig(-7.5) \big) \Big)
    \land \dots \\
    \dots \land 
    \Big( x \in \big[ 7.5, 8.0\big) \,\to\, y \in \big[ \sig(7.5), \sig(8)\big) \Big)
    \land \Big( x\in \big[8, +\infty\big) \,\to\, y \in \big[ \sig(8), 1\big] \Big).
  \end{gather*}
  The intention behind this approach was to find out whether such
  an approximation could have a positive effect on the running times of iSAT.
\end{enumerate}

\paragraph{Dedicated Propagator}
We shortly discuss the theoretical background of the sigmoid propagator and
describe how we implemented this propagator into iSAT.  For the general approach
to implement arbitrary propagators into iSAT, the interested reader is referred
to \cite[Ch. 5]{herde}.  Let $f: \mathbb{R} \to \mathbb{R}$ be a %an unary real
function and $\mathbb{I}$ be the set of real-valued intervals.  Further, let $y=
f(x)$ be an equation relating two variables $x \in X$ and $y \in Y$ with $X,Y\in
\mathbb{I}$. An \emph{interval constraint propagator} $\rho_{f}: \mathbb{I}^2
\to \mathbb{I}^2$ for $f$ maps $X \times Y$ to the smallest subset $X' \times
Y'$ in $X \times Y$ that contains all solutions to ${y = f(x)}$ that have been
in $X \times Y$, i.e.\ $\forall (x,y) \models y = f(x)$ we have $(x,y) \in X
\times Y \implies (x,y) \in X' \times Y'$. Such a propagator thus chops parts
from $X$ and $Y$ for which no solution to $y = f(x)$ exists, cf. 
\autoref{fig:sig}.%{\todopk{What are the publisher-rules for these abbreviations?
%It's sometimes to use ``Figure'' when starting a sentence with the word.}}
%depicts an example of an interval constraint propagation for $f(x) = \sig(x)$.

Since $\sig(x)$ is continuous and strictly monotonic increasing, the
boundary values of $X'$ and $Y'$ can be deduced directly from the boundary
values of $X$ and $Y$ by applying $\sig(x)$ and its inverse
$\sig^{-1}(x)$. The cases that $X$ is not bounded or that the bounds of $Y$
are not in the domain of $\sig(x)$ are special.
In order to cover these cases, we lift $\sig(x)$ and $\sig^{-1}(x)$ to
the extended field of real numbers $\overline{\mathbb{R}} = \mathbb{R} \cup
\{{-\infty, \infty}\}$ by defining two functions
$\sigma: \overline{\mathbb R} \to \overline{\mathbb R}$ and
$\sigma^{-1}: \overline{\mathbb R} \to \overline{\mathbb R}$ as
\begin{gather*}
    \begin{aligned}
        \sigma(x) := \begin{cases}
            0     & \text{ if } x = {-\infty} \\
            \frac{1}{1 + \mathrm e^{-x}} & \text{ if } {-\infty} < x < \infty \\
            1     & \text{ if } x = +\infty,\\
        \end{cases}
    \end{aligned}\qquad
    \begin{aligned}
        \sigma^{-1}(y)  := \begin{cases}
            -\infty     & \text{ if } y \leq 0 \\
            -\ln(\frac{1}{y}-1))  & \text{ if } 0 < y < 1\\
            +\infty     & \text{ if } 1 \leq y.
        \end{cases}
    \end{aligned}
\end{gather*}

The interval constraint propagation for the sigmoid function is implemented in
iSAT as two functions representing the so-called \emph{forward propagation} that
yields $Y'$, and the \emph{backward propagation} that yields $X'$. The
corresponding algorithms $\mathsf{fwd\_prop}_\sigma(X,Y)$ and
$\mathsf{bwd\_prop}_\sigma(X,Y)$ are listed in Alg.~\ref{alg:fwdprop}
and Alg.~\ref{alg:bwdprop} where $({\bowtie}_1, x_1, {\bowtie}_2, x_2 )$ with $x_1,
x_2 \in\overline{\mathbb R}$ and ${\bowtie}_1, {\bowtie}_2 \in \{{<},{\leq}\}$
represents an interval $I = \{x \mid x_{1} \bowtie_{1} x \bowtie_{2} x_{2}\}$.
Please note that the floor and ceiling notations $\lfloor\,\rfloor$ and
$\lceil\,\rceil$ here refer to a rounding towards the next floating point number
available on the computer on which iSAT is executed. Rounding of boundary values
is required since the finite precision of machine-representable numbers does not
allow to represent arbitrary real values.  However, such \emph{safe outward
rounding} guarantees that all solutions to $y = \sig(x)$ from $X \times Y$ are
included in $X' \times Y'$.

\begin{figure}
    \begin{minipage}[t]{0.45\textwidth}
        \begin{algorithm}[H]
            \KwIn{ $X = ( {\bowtie}_1, x_1, {\bowtie}_2, x_2)$, $Y$ }
            $y_1 := \lfloor\sigma(x_1)\rfloor$\;
            $y_2 := \lceil\sigma(x_2)\rceil$\;
            ${\bowtie}_1' := {\bowtie}_1$\;
            ${\bowtie}_2' := {\bowtie}_2$\;
            \lIf{ $y_1 = -\infty$ }{${\bowtie}_1' := {<}$}
            \lIf{ $y_2 = +\infty$ }{${\bowtie}_2' := {<}$}
            $Y' := Y \cap (\bowtie_1',y_1,\bowtie_2',y_2)$\;
            return $Y'$\;
            \caption{$\mathsf{fwd\_prop}_\sigma(X,Y)$}
            \label{alg:fwdprop}
        \end{algorithm}
    \end{minipage}
    \hfill
    \begin{minipage}[t]{0.45\textwidth}
        \begin{algorithm}[H]
            \KwIn{$X$, $Y = ( {\bowtie}_1, y_1, {\bowtie}_2, y_2)$}
            $x_1 := \lfloor\sigma^{-1}(y_1)\rfloor$\;
            $x_2 := \lceil\sigma^{-1}(y_2)\rceil$\;
            ${\bowtie}_1' := {\bowtie}_1$\;
            ${\bowtie}_2' := {\bowtie}_2$\;
            \lIf{ $x_1 = 0$ }{${\bowtie}_1' := {<}$}
            \lIf{ $x_2 = 1$ }{${\bowtie}_2' := {<}$}
            $X' := X \cap (\bowtie_1',x_1,\bowtie_2',x_2)$\;
            return $X'$\;
            \caption{$\mathsf{bwd\_prop}_\sigma(X,Y)$}
            \label{alg:bwdprop}
        \end{algorithm}
    \end{minipage}
\end{figure}

Such a dedicated propagator allows to encode a sigmoidal ANN directly into iSAT
without any paraphrasing through composition of functions. In addition, the
direct encoding involves only one operator per sigmoidal activation and thus a
single propagator call; the compositional approach instead requires multiple
propagator calls, namely one for each arithmetic function used in the encoding.
We thus expect the dedicated propagator to outperform the compositional
approach.

%% file: further_input/experiment_setup/setup.tex
\subsection{Experimental Setup}
\label{sec:experimentalsetup}

We describe the networks and applications we used to compare
the approaches. We trained various networks for two applications, namely 
\begin{enumerate*}[label=\alph*)]
    \item to give an emergency brake advisory within a train control
        system, and
    \item to recognise hand-written digits.
\end{enumerate*}

We performed two types of evaluation on these networks. Based on the train
control example, we compared the runtime performance of the dedicated propagator
to the compositional and the approximating approach, the results are presented in \autoref{sec:firstexperiments}. In \autoref{sec:verificationexperiments},
we aimed at a more severe verification task using dedicated safety
properties. We performed corresponding benchmarks based on the
digit recognition example as well as on the train control
example. For all these benchmarks, we translated ONNX
\cite{onnx} models of our trained networks into an
SMT formula in the iSAT language which then was extended by additional
constraints specifying the property to verify.

The remainder of this section describes the two applications, the networks we
trained for our benchmarks, and the properties we tried to verify.

\input{further_input/experiment_setup/etcs_intro}

\input{further_input/experiment_setup/mnist_intro}

%% file: further_input/experiment_setup/etcs_intro.tex
\subsubsection{European Train Control System (ETCS)}
\label{sec:etcsintro}

\begin{figure}[h]
    \centering
    \includegraphics[width=0.5\textwidth]{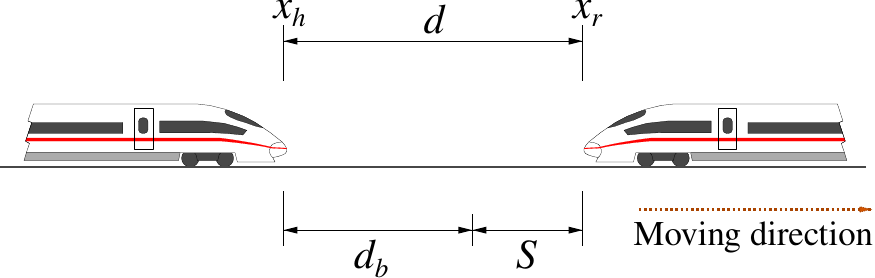}
    \caption{Simplified European Train Control System (ETCS) model}
    \label{fig:etcs_intro}
\end{figure}

One of the applications used for our evaluation is a simplified emergency brake
advisory function from the moving block authority of the European Train Control
System according to \cite{herde}: Two trains are running behind each other on
one track into the same direction (see \autoref{fig:etcs_intro}). The train
behind shall be advised to initiate an emergency braking manoeuvre if it cannot
stop without violating a safety distance to the train ahead, i.e.\ the distance
to the train ahead is smaller than the minimum braking distance (using maximum
deceleration) plus a safety distance. We consider the worst case scenario where
the train ahead is at standstill.

We assume that the maximum deceleration for the train behind is
$\SI{-0.7}{\metre\per\square\second}$ and a safety distance
$S=\SI{400}{\metre}$. Let $x_h$ describe the head position of the train behind,
while $x_r$ is the rear position of the train ahead. The maximum braking
distance $d_{b}$ available before entering the safety margin to the train ahead
and the deceleration $a$ required to stop with exactly the safety distance are calculated as
\[
    d_{b} = x_r - (x_h + S)  \text{\hspace{1cm}and\hspace{1cm}}
    a = -\frac{v^{2}}{2d_{b}}\;,
\]
where $v$ is the velocity of the train behind. 
The system shall then advise for an emergency brake if and only if
the deceleration required to stop with exactly the safety margin is larger than
the maximum deceleration, i.e.\ 
\begin{align*}
    \mathit{braking} \equiv \begin{cases}
        \text{true}  & \text{if } a < \SI{-0.7}{\metre\per\square\second}\;,  \\
        \text{false}   & \text{otherwise.}
    \end{cases}
\end{align*}

We used the Keras API~\cite{keras} in combination with the Adam optimisation
algorithm~\cite{Adam} to train 16 fully connected feedforward ANNs for this
function, each of which varying in the number of layers, the number of neurons
per layer, and the total number of neurons as illustrated in \autoref{tab:etcs}.
Each network has three input neurons for $v$, $x_h$, and $x_r$ as well as two output
neurons $\outp_{0}$ and $\outp_{1}$.
The output value of $\outp_0$ votes for ``not braking'' and the value of $\outp_1$ votes for ``braking''. The network indicates the braking advisory $\mathit{braking} \equiv \mathrm{true}$
if and only if the vote of $\outp_1$ outweighs the vote of $\outp_0$, i.e. if and only if $\outp_{0} \leq \outp_{1}$.

The training was performed based on a self-generated data set containing
$200\,000$ entries. Each entry contains the velocity $v$ of the train behind,
the track positions $x_r$ and $x_h$, and the braking advisory
$\mathit{braking}$.
The velocity may range from $0$ to $\SI{83.4}{\metre\per\second}$, the
length of the track is $\SI{50000}{\metre}$, and the safety distance is not
violated trivially, i.e.\ we always have $d_{b} \geq 0$. 

\begin{table}
  \centering
    \caption{Distribution of neurons in ETCS and MNIST networks
        (fully connected and feedforward). The neurons per hidden layer are
        explicitly specified per layer for the MNIST networks.}
    \label{tab:etcs}
  \footnotesize
  \begin{tabular}{@{}l  rr  rrr@{}}
    \toprule
    {\bf model name}
    & \multicolumn{2}{c}{\bf input / output layer}
    & \multicolumn{3}{c@{}}{\bf hidden layers}\\
    & \multicolumn{2}{c}{$\overbrace{\rule{3.2cm}{0pt}}$}
    & \multicolumn{3}{c@{}}{$\overbrace{\rule{6.6cm}{0pt}}$}\\    
    %\midrule
    & input nodes & output nodes 
    & no.\ of layers & neurons per layer
    & total no.\ of neurons\\
    \midrule
    \textit{ETCS} & 3 & 2 & 2 & 12 & 24\\
    & 3 & 2 & 2 & 25 & 50\\
    & 3 & 2 & 2 & 50 & 100\\
    & 3 & 2 & 2 & 100 & 200\\
    & 3 & 2 & 3 & 12 & 36\\
    & 3 & 2 & 3 & 25 & 75\\
    & 3 & 2 & 3 & 50 & 150\\
    & 3 & 2 & 3 & 100 & 300\\
    & 3 & 2 & 4 & 12 & 48\\
    & 3 & 2 & 4 & 25 & 100\\
    & 3 & 2 & 4 & 50 & 200\\
    & 3 & 2 & 4 & 100 & 400\\
    & 3 & 2 & 5 & 12 & 60\\
    & 3 & 2 & 5 & 25 & 125\\
    & 3 & 2 & 5 & 50 & 250\\
    & 3 & 2 & 5 & 100 & 500\\\hline
    \textit{MNIST} & 784 & 10 & 2 & (392, 196) & 588\\
    & 784 & 10 & 2 & (784, 392) & 1\,176 \\
    \bottomrule
  \end{tabular}
\end{table}

\paragraph*{Properties to Verify}

We used these networks for both types of evaluation: for the comparison of
approaches and for the dedicated attempt to verify a property.

For the comparison of approaches, we used several verification targets, some of
them expected to yield a satisfiable constraint system as well as some expected
to yield an unsatisfiable constraint system, thereby aiming at a broad
overview of how the approaches perform on the class of constraint systems
yielded by our encoding approaches. Throughout all verification tasks, we
investigated whether the advice $\mathit{braking} \equiv \mathrm{true}$, indicated by $\outp_0 \le \outp_1$,
is satisfiable or not, i.e.\ we added a constraint $\outp_{0} > \outp_{1}$
to the iSAT input s.t.\ the desired property is satisfied iff the resulting
constraint system is unsatisfiable. The
verification tasks varied in the search space, i.e.\ in their constraints
on the input to the ANN, as in the following scenarios:
\begin{enumerate}[label={(\Alph*})]
    \item We left the inputs unconstrained.
    \item We restricted positions to $x_r > x_h$, i.e.\ to
        situations where there train behind runs with some positive distance
        behind the train ahead.
    \item We restricted the train behind to a velocity $v >
        \SI{25}{\metre\per\second}$ and the position $x_h =
        \SI{15000}{\metre}$ and the position of the train ahead was restricted
        to $x_r = \SI{35000}{\metre}$. In this scenario, the property is
        expected not to be satisfied.
    \item\label{scen:unsat} Again, we restricted the train behind to a velocity $v >
        \SI{25}{\metre\per\second}$ while the positions of both trains were
        constrained to ${x_h, x_r < \SI{800}{\metre}}$. This
        yields only situations where violating the safety distance $S$ is unavoidable due to the
        limited maximum distance between the trains. Thus, the property is
        expected to be unsatisfied.\label{enum:target3}
\end{enumerate}
We applied all these variants to all ETCS networks listed in \autoref{tab:etcs}
for all three approaches, namely the compositional approach, the approximating
approach, and the dedicated propagator.

When aiming at a severe verification task, we assumed that the velocity of
the train behind is ${v \in (\SI{20}{\metre\per\second},
\SI{80}{\metre\per\second}]}$, and that the distance $d =
x_r-x_h$ between the trains is smaller than the safety margin
$S$, i.e.\ $d \in [\SI0\metre,S]$. This setup inherently requires an advisory
$\mathit{braking} \equiv \mathrm{true}$. We hence added the corresponding
constraints on the inputs of the ANN under inspection to the iSAT input plus a
constraint $\outp_{0} > \outp_{1}$ and aimed at a proof that the resulting
constraint system is unsatisfiable, i.e.\ that the corresponding ANN always
advised for an emergency brake when the distance between the trains is smaller
than the safety margin.

%% file: further_input/experiment_setup/mnist_intro.tex
\subsubsection{Recognition of Handwritten Digits}
\label{sec:mnistintro}

We trained two fully connected feedforward networks for the recognition of
handwritten digits each of them comprising two hidden layers and varying in
the number of neurons per layer as well as the total number of neurons as indicated
in \autoref{tab:etcs}.

\begin{figure}[H]
	\centering
	\includegraphics[width=0.3\textwidth]{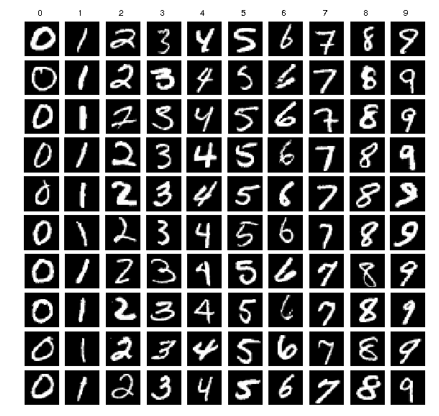}
	\caption{Handwritten digits from the MNIST data set }
	\label{fig:mnist_intro}
\end{figure}

The training was based on the MNIST data set \cite{mnist_dataset} which is 
widely used in the context of artificial neural network classification. The
data set consists of over 60.000 entries of grey-scaled images of size 28x28
pixels with each pixel value being an integer in $[0,255]$. Each image
represents a handwritten digit from 0-9 (see \autoref{fig:mnist_intro}).

\paragraph*{Property to Verify}

For this benchmark we drew a number of samples from the MNIST database.
We searched
for counterexamples violating a property to guarantee within an
$\varepsilon$-box around the sample, i.e.\ for any sample $s$ with vector
components $s_{k}$ and $1 \leq k \leq 28^2$, we constrained the inputs to the
network within the constraint system describing the network to $\max(0, s_{k} -
\varepsilon) \leq \inp_{k} \leq \min(1, s_{k} + \varepsilon)$ with $\varepsilon
= 2.55$.\footnote{This non-integer $\varepsilon$-value is an artefact of the
normalisation of input data; in fact, we used $\varepsilon = 0.01$ which is
$2.55$ in the non-normalised domain.}

Let $j$ be the true digit of the sample. For each sample, we formulate the
verification targets, i.e.\ the constraints added to the constraint system, as
follows. We asked for a unique identification of $j$, i.e.\ we tried to prove
that the output value for $\outp_{j}$ is the largest. A counterexample to this
property would thus satisfy the constraint $\bigvee_{i=0}^{9} \outp_{i} \geq
\outp_{j}$ for $i \neq j$.

In order to alleviate the solving process, we split up this condition along the
disjunction into multiple constraint systems instead of testing all outputs
within a single solver run. We thus created a constraint system for each $i \neq
j$ for each sample comprising a constraint $\outp_{i} \geq \outp_{j}$. We thus
moved the disjunction of the above condition to file level, i.e.\ the property
is not satisfied if one of those constraint systems is satisfiable. For this
benchmark, we drew 20 samples, thereby yielding 180 constraint systems to solve
for both digit recognition networks.

%% file: further_input/feasibility.tex
\subsection{Comparison of Approaches}
\label{sec:firstexperiments}

\begin{figure}
    \begin{subfigure}[c]{0.49\textwidth}
        \includegraphics[width=\textwidth]{%
            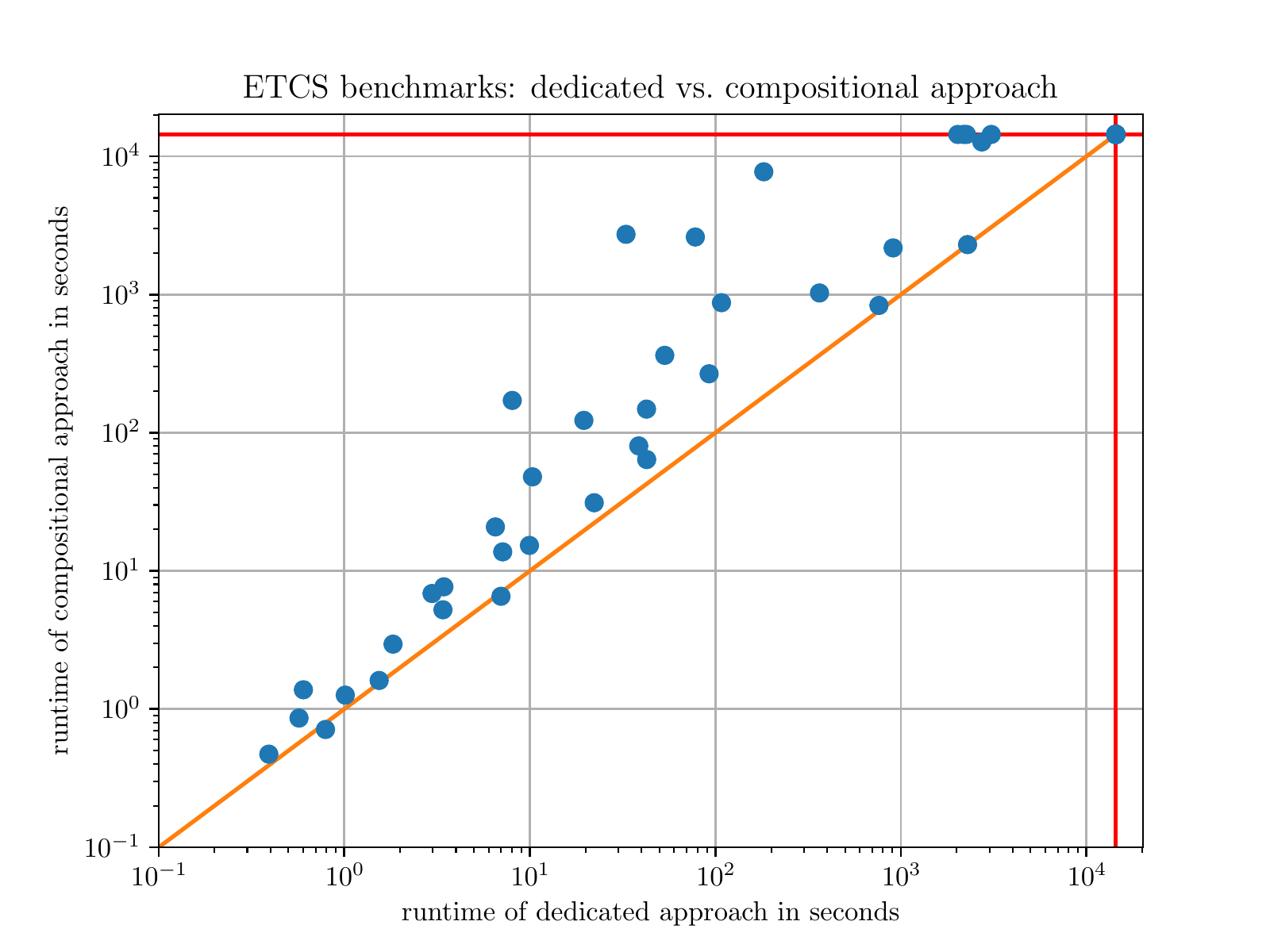}  
        \subcaption{Dedicated propagator vs.\ compositional approach}
        \label{fig:runtime1-composed}
    \end{subfigure}\hfill
    \begin{subfigure}[c]{0.49\textwidth}
        \includegraphics[width=\textwidth]{%
            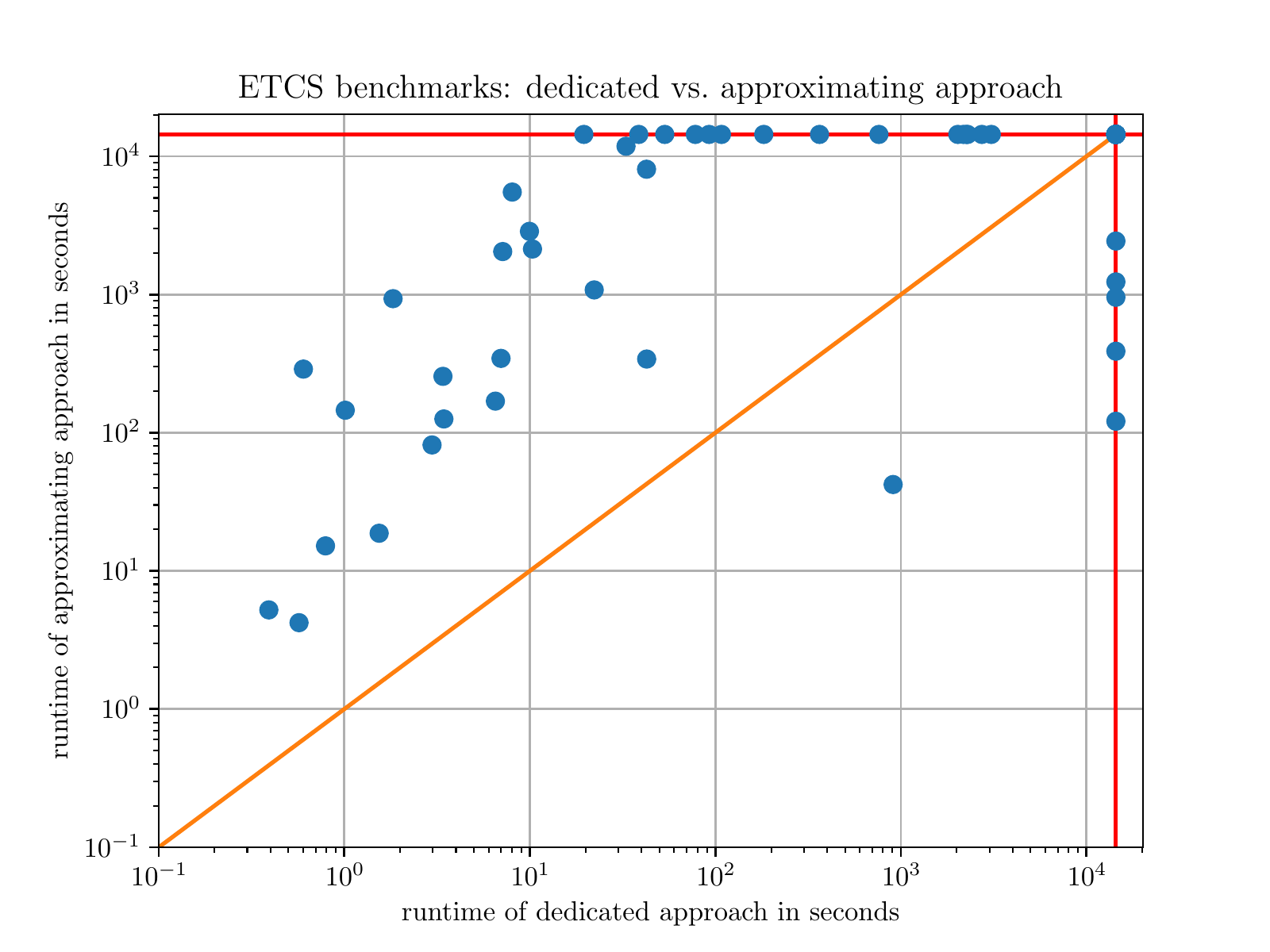}  
        \subcaption{Dedicated propagator vs.\ approximating approach}
        \label{fig:runtime1-approx}
    \end{subfigure}
    \caption{Comparison of the runtimes of the approach using the dedicated
    sigmoid propagator with the runtimes of the compositional approach 
    (a) and the approximating approach (b). The horizontal axes show the
    runtimes using the dedicated propagator while vertical axes show the
    runtimes of the comparative approaches (all axes are in logscale). Each
    data point illustrates the runtimes of the compared approaches on a
    particular network and verification target. Timeouts are located on the red
    borders. The orange line indicates the line of equal runtime. The benchmarks
    were performed with a timeout of 4 hours (14,400 seconds).}
    \label{fig:runtimecomb}
\end{figure}

We compare the dedicated sigmoid propagator with the compositional approach and
the approximating approach based on runtime evaluations. To this end, we
executed benchmarks on the ETCS networks described in \autoref{tab:etcs} where
we investigated whether the advice $\mathit{braking} \equiv \mathrm{true}$
is satisfiable or not in four different settings (see \autoref{sec:etcsintro}). The
maximum runtime was set to $4$ hours for each solver run.

\Autoref{fig:runtime1-composed} illustrates the comparison with the
compositional approach and indicates that the dedicated sigmoid propagator
performs not worse than the compositional approach throughout all networks and
verification tasks. However, the performance of the dedicated propagator seems
to be noticeable better than that of the compositional approach when focusing on
large runtimes (please note the logarithmic scales). This advantage of the
dedicated propagator is also reflected in the absolute number of timeouts: the
dedicated approach terminated on $36$ of the $64$ solver runs within $4$ hours
(that is a timeout rate of $43.75\%$), while the compositional approach
terminated on $32$ runs (timeout rate $50\%$). The overall runtime spent for the
dedicated approach\footnote{Timeouts account for their actual runtime, i.e.\ $4$
hours. The overall runtime thus provides a lower bound on the runtime required
for all solver runs on the corresponding benchmark set.} is $\approx 4.87$ days
($\approx 1.83$ hours on per run on average).  The compositional approach, in
contrast, required an overall runtime of $\approx 5.74$ days ($\approx 2.15$
hours on per run on average).

A more clear difference between the dedicated and the approximating approach is
illustrated in \autoref{fig:runtime1-approx}: Here, for many solver runs, the
runtime of the approximating approach is roughly about one decimal power or more
larger than for the dedicated approach.
There are also runs on which the approximating approach
is much faster; however, this approach terminated on $26$ runs only (that is a
timeout rate of $59.38\%$) and required an overall runtime of $\approx 7.14$
days ($\approx 2.67$ hours on per run on average). For the other two approaches, iSAT was able to prove the desired property for the networks with 2 layers with 25 and 50 nodes respectively while the
approximating approach was not able to prove that the desired property is
satisfied in the ETCS-scenario \ref{enum:target3} %with the unevitable collision
for any of the $16$ networks. In particular for those two networks for which the other approaches
could prove the property, iSAT terminated with a less informative candidate
solution in one case, and was aborted due to timeout in the other case.

We speculate that the Boolean complexity of the verification problem increases
strongly due to the piecewise definition of the approximation formula, and hence
leads to a considerable increase of the runtime. A further investigation of
whether this assumption is true is still pending and will be addressed as future
work.  In view of the disappointing performance of the approximating approach, we
do not pursue it in the following.

%% file: further_input/prep_patch.tex
\subsection{Improvements in Preprocessing}
\label{sec:preprocessing}

\begin{figure}
    \begin{subfigure}{0.45\linewidth}
        \begin{center}
            \includegraphics[width=\textwidth]{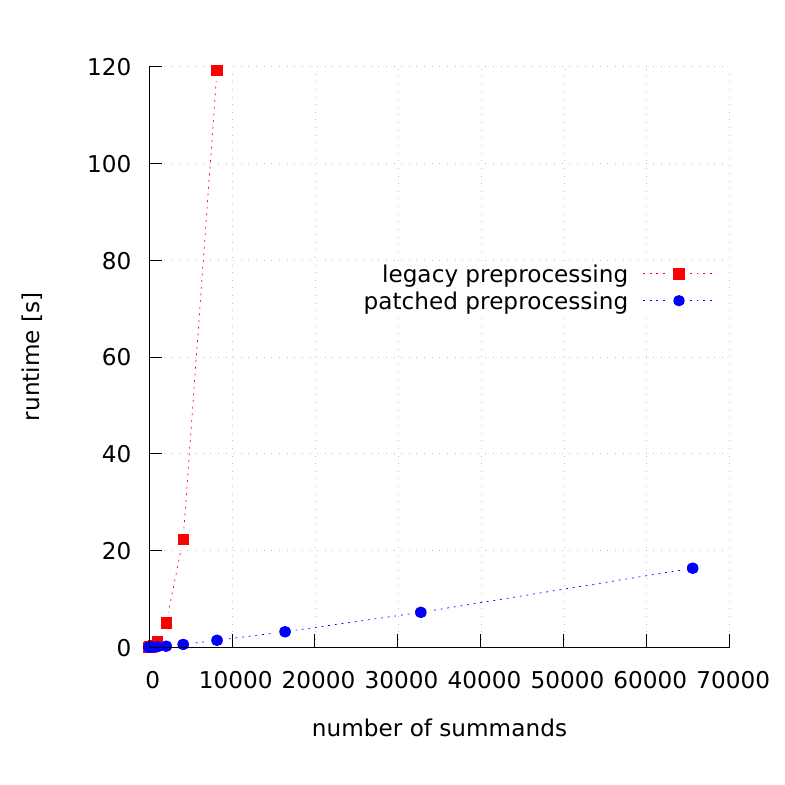}
            \caption{Runtime of preprocessing.}\label{fig:prepro benchmark}
        \end{center}
    \end{subfigure}
    \hfill
    \begin{subfigure}{0.45\linewidth}
        \begin{center}
            \includegraphics[width=\textwidth]{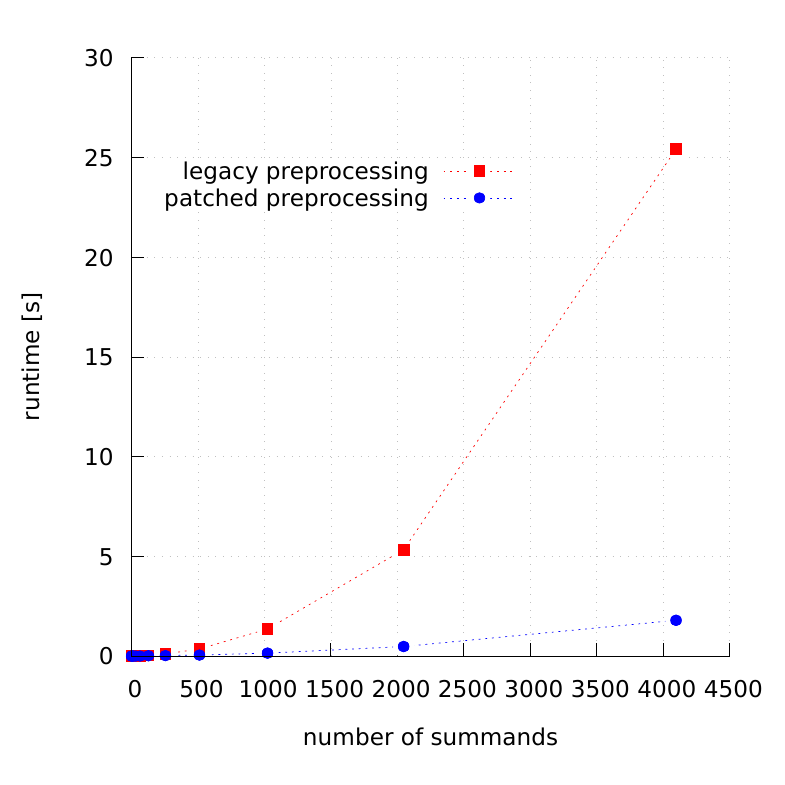}
            \caption{Runtime of solving.}\label{fig:solving benchmark}
        \end{center}
    \end{subfigure}
    \caption{Benchmarks for $\mathsf{sum}_n$. The timeout was set to 300 sec and
    the memory limit was 16GB.}
\end{figure}

During our experiments we observed that iSAT spends a lot of time in the formula
preprocessing steps\,---\, in particular for models involving summations over
many terms as they naturally occur in vector and matrix operations.
To this end, we used a simple scalable benchmark $\mathsf{sum}_n$ asking for
a satisfying solution of the equality $y = \sum_{i=0}^n \tfrac{1}{n}x_i$
with $y,x_i\in[0,1]$ for $i=0,\dots,n$ to measure the runtime of iSAT's
preprocessing. We observed an exponential relationship between the problem
size $n$ and the runtime of the preprocessing, see the
plot 'legacy preprocessing' in \autoref{fig:prepro benchmark}.

In a deeper analysis, we found that iSAT tends to build an unbalanced abstract
syntax tree during parsing and normalisation steps. Hence, recursions over
the syntax tree often suffers from the unnecessary depth of the syntax tree.
We patched the preprocessing steps of iSAT so that iSAT now builds
well-balanced summation terms of minimal depth.
Interestingly, this approach also helped to reduce the solving time of iSAT as
depicted in \autoref{fig:solving benchmark}. Both plots in
\autoref{fig:solving benchmark} still behave exponentially.
However, it is noticeable that the plot for the patched version is much flatter
compared to the legacy version. Note that in both versions
the benchmarks for $n>4096$ yield a memout and, hence, are not depicted
in the plots.

A deeper examination of the reasons why also the solving time profits from
well-balanced trees and whether these benefits can be extended to other
operators is considered as an interesting future work.

\FloatBarrier

%% file: further_input/new_benchm.tex
\subsection{Towards Verification of Safety Properties}
\label{sec:verificationexperiments}

\begin{figure}
    \begin{subfigure}[c]{0.49\textwidth}
        \includegraphics[width=\textwidth]{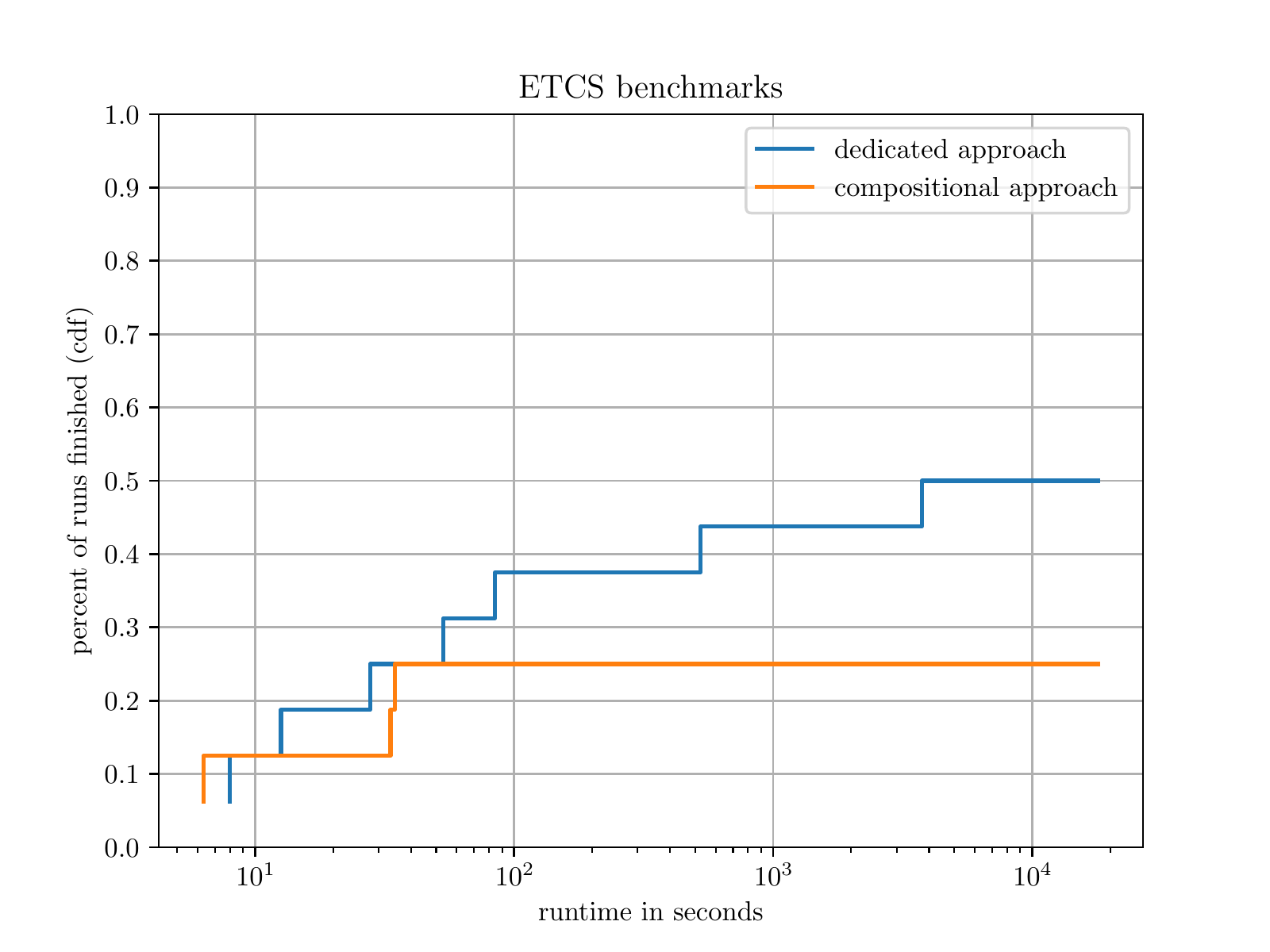}  
        \subcaption{Benchmark results for the dedicated and the composed
        propagator on the more severe verification ETCS verification task
        from \autoref{sec:etcsintro}.}
        \label{fig:etcs2result}
    \end{subfigure}\hfill
    \begin{subfigure}[c]{0.49\textwidth}
        \includegraphics[width=\textwidth]{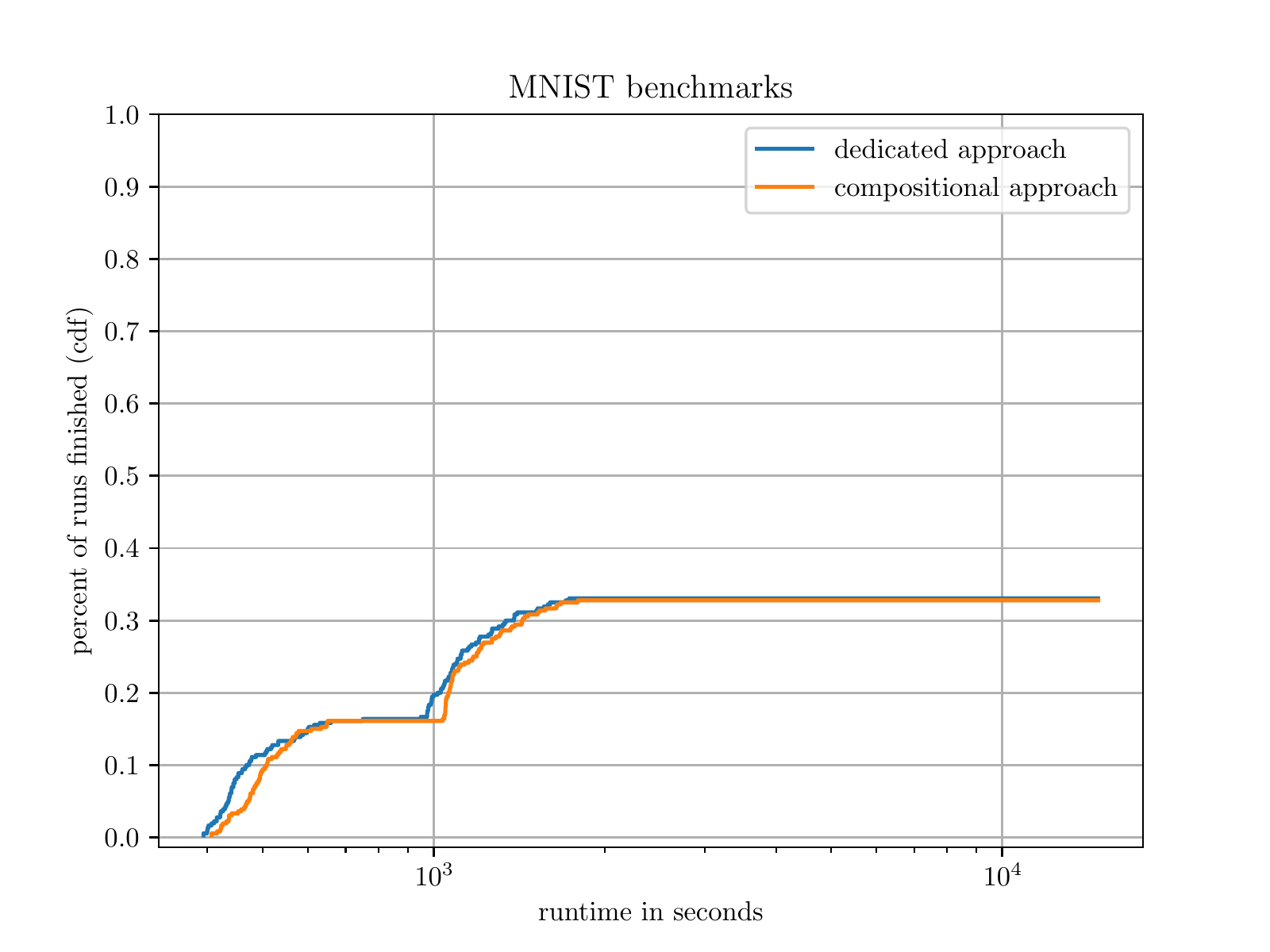}
        \subcaption{Benchmark results for the dedicated propagator on
        on the MNIST verification tasks described in
        \autoref{sec:mnistintro}.\\\phantom{nasty fix for caption height}}
        \label{fig:mnistresult}
    \end{subfigure}
    \caption{Benchmark results on the somewhat more severe verification
    tasks. The plots show the proportion of constraint systems that have been
    solved over time (cumulative distribution function).
    The benchmarks were performed with the patched iSAT version with a timeout of 4
hours (14,400 seconds).
}
    \label{fig:benchmark2a}
\end{figure}

\begin{figure}
	\begin{subfigure}[c]{0.49\textwidth}
		\includegraphics[width=\textwidth]{%
			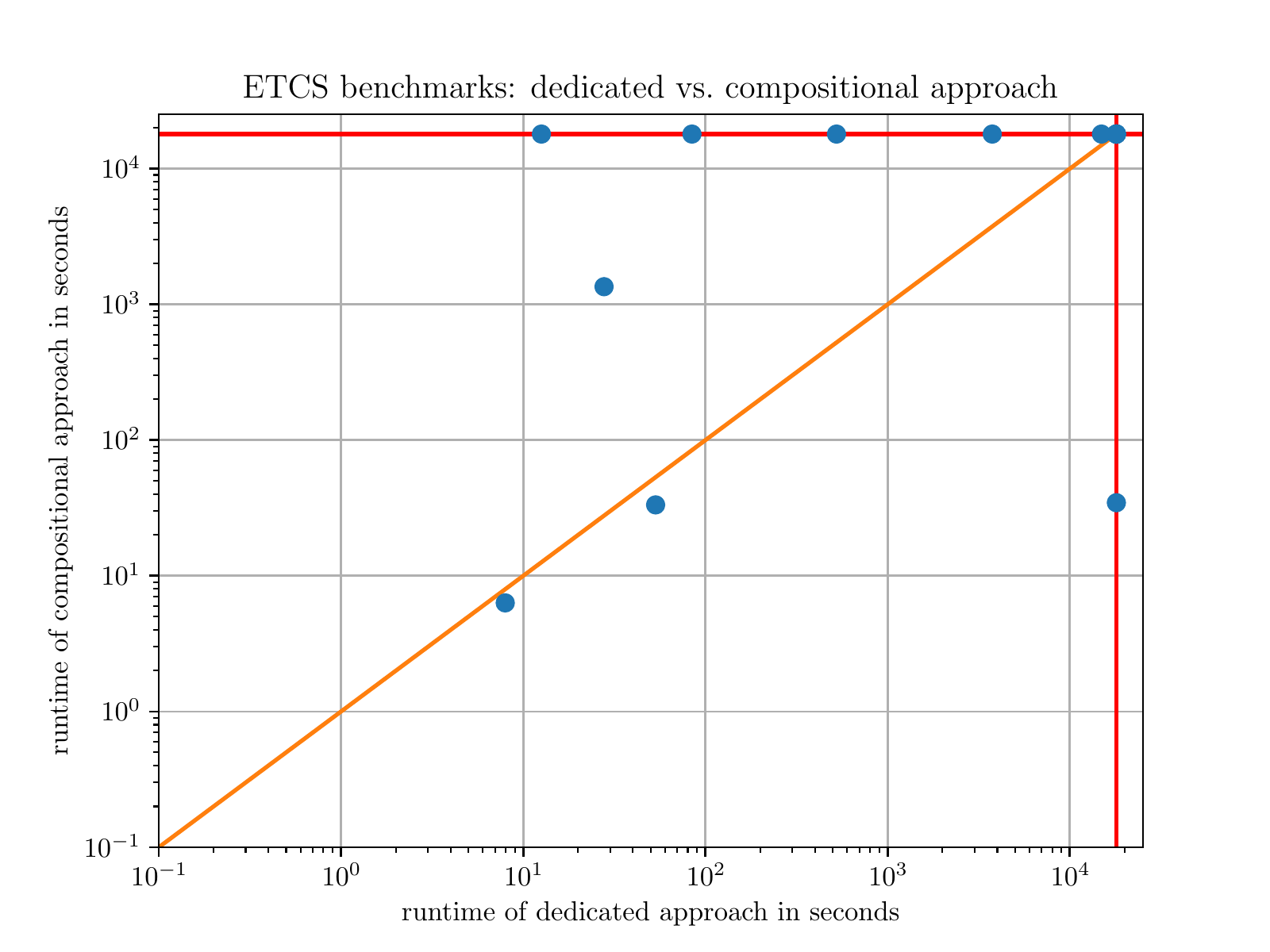}  
		\subcaption{ETCS benchmark}
		\label{fig:runtime2-composed}
	\end{subfigure}\hfill
	\begin{subfigure}[c]{0.49\textwidth}
		\includegraphics[width=\textwidth]{%
			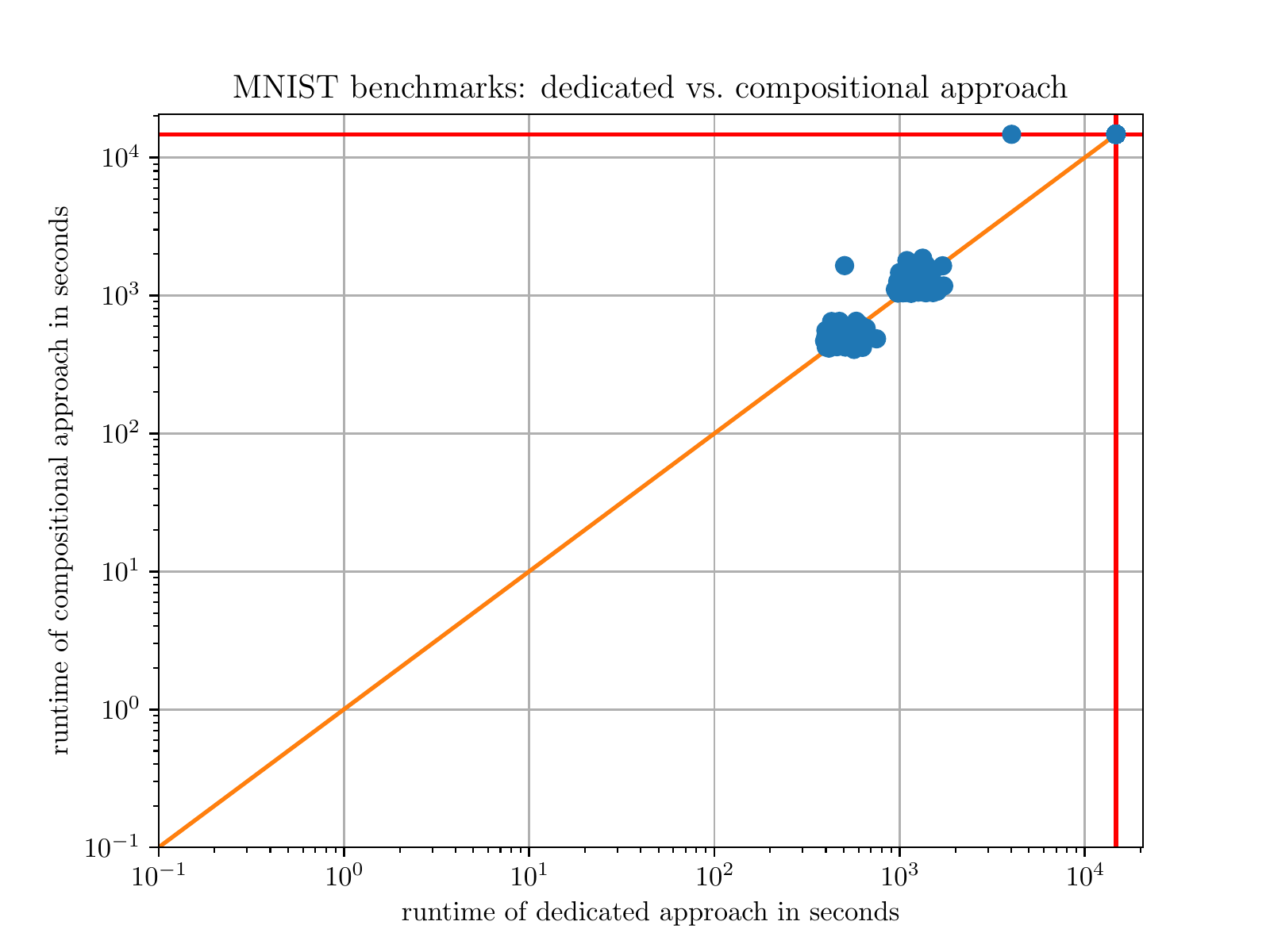}  
		\subcaption{MNIST benchmark}
		\label{fig:runtime2-mnist}
	\end{subfigure}
	\caption{Comparison of the runtimes of the dedicated and the compositional
		approach on  
		(a) the more severe verification task on the ETCS networks described in
		\autoref{sec:etcsintro} and (b) the verification task on digit recognition
		networks described in \autoref{sec:mnistintro}. The horizontal axes show the
		runtimes of the dedicated approach while the vertical axes show the
		runtimes of the compositional approach (all axes are in logscale). Each
		data point illustrates the runtimes of the compared approaches on a
		particular network. Timeouts are located on the red borders. The orange line
		indicates the line of equal runtime. The benchmarks were performed with the
		patched iSAT version with a timeout of 4 hours (14,400 seconds).}
	\label{fig:benchmark2b}
\end{figure}

According to our previous results, in the following we pursued only the
compositional and the dedicated approach. Here, we performed benchmarks that
were less broadly based but examined somewhat more severe verification tasks, as
a successful verification of a safety property corresponds to unsatisfiability
result for the negation of the property under investigation. Hence, iSAT has to
exclude the presence of (candidate) solutions which is\,---\,in general\,---\,a
more demanding task.

We employed two different benchmark settings to this end, firstly the last ETCS
setting from \autoref{sec:etcsintro} in which we try to verify that a network
always gives an advisory $\mathit{braking} \equiv \mathrm{true}$ whenever the
safety distance is trivially violated, and secondly the MNIST scenario where we
try to prove the absence of adversarial images within some small
$\varepsilon$-box around images we sampled from the MNIST database as described
in \autoref{sec:mnistintro}.

The result for the ETCS setting are illustrated in \autoref{fig:etcs2result}.
The figure shows cumulative distribution functions, i.e.\ the proportion of
terminated solver runs plotted over the runtime. Here, we can clearly identify
that the dedicated approach can solve more constraint systems ($8$ out of $16$,
i.e.\ $50\%$) within the time limit than the compositional approach ($25\%$).
The compositional approach was able to prove two networks to be safe, i.e.\ iSAT
terminated with ``unsatisfiable'' for the networks with $12$ nodes per layer and
$2$ and $5$ layers respectively. These networks were also proven to be safe by
the dedicated approach, plus $4$ additional networks ($2$ layers with $50$ and
$100$ nodes per layer, $3$ layers with $12$ nodes per layer, and $4$ layers with
$50$ nodes per layer) for which the compositional approach was aborted due to
timeout.

A somewhat different result is shown for the MNIST benchmark in
\autoref{fig:mnistresult}: No clear distinction between the two approaches is
possible wrt.\ the cumulative distribution function. We cannot even determine,
apart from outliers, that there are constraint systems for which one or the
other approach is faster and that these differences balance out in total.
\Autoref{fig:runtime2-mnist} reveals that\,---\,apart from outliers\,---\,the
runtime is essentially equal for almost all constraint systems. The absolute
numbers also do not exhibit significant differences: The dedicated approach was
aborted due to timeout in $241$ cases ($66.94\%$ timeout rate) and proved $119$
networks to be safe ($33.06\%$) while the compositional approach was aborted in
$242$ cases ($67.22\%$ timeout rate) and proved $118$ networks to be safe
($32.78\%$). Successful verification of the desired property and timeouts were distributed almost equally over
both network sizes for both approaches.

We thus cannot clearly conclude from these results of our second evaluation
step, that the dedicated approach is always to be preferred over the
compositional approach when aiming at severe verification
tasks.

%% file: further_input/discussion.tex
\section{Conclusion}
\label{sec:conclusions}
The aim of our paper was to investigate to what extent iSAT, an SMT solver
that inherently supports nonlinear and transcendental functions, is suitable
for the automatic verification of severe safety properties of ANNs.
To this end, we first considered three different encoding approaches
for the functional relation of an ANN. While the layer-wise weighted summation
over input neurons can be encoded as linear combinations, these approaches
differ in their handling of the sigmoid activation function.
In the approximating approach the sigmoid activation function is replaced by
encapsulating interval boxes which is inherently inexact. The compositional
and the dedicated approach fully rely on iSAT's ability to
handle nonlinear and
transcendental functions, whereas the dedicated approach required the
implementation of a sigmoid propagator into iSAT explicitly.
In our preliminary investigation, we noticed that the preprocessing of iSAT
shows some weaknesses with regard to the treatment of extensive
linear combinations, which we were able to eliminate by a patch.
Thereafter, we could finally turn to the verification of severe
safety properties.

Through our investigation, we were able to determine that it is not
worthwhile to rely on the approximating approach, neither in terms of runtime
nor the quality of the results. It is more advisable to rely on the strengths
of iSAT's nonlinear and transcendental propagators, either following
the compositional or the dedicated approach.
While our preliminary investigation and also
the verification of safety properties for the ETCS example
suggest a rather clear preference for the dedicated approach,
we could not fully confirm this impression for the verification of 
safety properties in the MNIST example.

A preliminary conclusion is that the dedicated propagator can be useful,
but compared to the compositional approach, one must critically ask oneself
whether the implementation effort is justified. We can therefore recommend
the implementation of a dedicated propagator only for activation functions
that occurs sufficiently often in the networks to be examined.

Another important observation is that the solving time of iSAT
strongly depends on the network structure, its size,
and the verification targets.
Regarding the network structure, a closer look at the internal processes
of iSAT seems to be useful, as our experiences with the preprocessing issues
already revealed. For example, we could not conclusively clarify
what role the increased Boolean complexity of the approximating approach played.
Especially with regard to the comparison of the dedicated and
compositional approach, a deeper investigation of iSAT's splitting mechanism
and its various heuristics seems to be important. It is not clear whether
the additional splitting possibilities on the auxiliary variables that are
necessary for the compositional approach are sometimes beneficial.
However, during our experiments, we tried different splitting heuristics
without getting a clear prioritisation of one of these heuristics.

Despite all these imponderables, we have nevertheless seen that iSAT
is able to verify networks with nontrivial nonlinear activation functions.
We see the strength of iSAT here not so much in the treatment of the
linear components, but rather in the ability to quickly support a
large variety of nonlinear activation functions through a
compositional approach, as well as in the possibility to extend iSAT
by dedicated propagators when a performance improvement is expected
due to such an extension.

As future work, we plan to further investigate the relationship of
the network structure and the verification targets with iSAT's
internal processes with the aim to identify those settings where
implementing a dedicated propagator pays off.
On the other hand, our results also suggest to extend our translation
toolchain to the point where a wide variety of activation functions
is supported by the compositional approach fully automatically.
Looking at the rapidly growing number of verification tools, a comparison of
runtime and precision of verdicts with other tools is also indispensable.